\documentclass[
]{ceurart}

\usepackage{listings}
\usepackage{graphicx}
\usepackage{amsmath}
\usepackage{algorithmic}
\usepackage{algorithm}
\begin{document}

\copyrightyear{2025}
\copyrightclause{Copyright for this paper by its authors.
  Use permitted under Creative Commons License Attribution 4.0
  International (CC BY 4.0).}

\conference{NORA'25: 1\textsuperscript{st} Workshop on Knowledge Graphs \& Agentic Systems Interplay co-located with NeurIPS, Dec.1, 2025, Mexico City, Mexico}

\title{Graph Distance as Surprise: Free Energy Minimization in Knowledge Graph Reasoning}

\author[1]{Gaganpreet Jhajj}[%
orcid=0000-0001-5817-0297,
email=gjhajj1@learn.athabascau.ca,
]
\cormark[1]
\address[1]{School of Computing and Information Systems, Athabasca University, Canada}

\author[1]{Fuhua Lin}[%
orcid=0000-0002-5876-093X,
email=oscarl@athabascau.ca,
]

\begin{abstract}
In this work, we propose that reasoning in knowledge graph (KG) networks can be guided by surprise minimization. Entities that are close in graph distance will have lower surprise than those farther apart. This connects the Free Energy Principle (FEP) \cite{friston_free-energy_2010} from neuroscience to KG systems, where the KG serves as the agent's generative model. We formalize surprise using the shortest-path distance in directed graphs and provide a framework for KG-based agents. Graph distance appears in graph neural networks as message passing depth and in model-based reinforcement learning as world model trajectories. This work-in-progress study explores whether distance-based surprise can extend recent work showing that syntax minimizes surprise and free energy via tree structures \cite{murphy_natural_2024}. 
\end{abstract}


\maketitle

\section{Introduction}

The Free Energy Principle (FEP) suggests that biological systems minimize surprise by maintaining accurate world models \citep{friston_free-energy_2010,friston_active_2017, parr_active_2022}. Recently, Murphy et al. \cite{murphy_natural_2024} demonstrated that syntactic operations minimize surprise through shallow tree structures. They quantified surprise via tree depth (geometric complexity) and Kolmogorov complexity (algorithmic complexity), approximated through Lempel-Ziv compression \citep{li_introduction_2008,ziv_universal_1977}.

In FEP, agents minimize variational free energy $F = -\log P(o,s) - H[Q(s)]$, where $o$ represents observations, $s$ hidden states, $P$ the generative model, and $Q$ the agent's beliefs \citep{friston_free-energy_2010}. The first term, $-\log P(o,s)$, quantifies surprise: entities with high probability under the generative model (high $P(o,s)$) yield low surprise (low $-\log P(o,s)$). For syntactic trees, Murphy et al. \cite{murphy_natural_2024} used tree depth to proxy this probability; we extend this principle to general graphs using shortest-path distance.

In active inference, minimizing free energy drives both perception (updating beliefs $Q(s)$) and action (selecting policies that reduce uncertainty) \citep{friston_active_2017}. We apply this principle to KG reasoning: entities at shorter graph distances have a higher probability under the agent's graph-based generative model. The central question we address is: given a KG serving as an agent's generative model, which entity groundings are plausible for a query in context? We propose one principled approach: plausibility inversely correlates with graph distance.

Knowledge graphs (KGs) are increasingly integrated with modern AI agents, with the ability to improve reasoning, memory, and planning \citep{chen_plan--graph:_2025, cui_prompt-based_2025, he_g-retriever:_2024,jhajj_educational_2024, jhajj2025neuromorphic, jhajj2024jack,morland2025adaptable,gustafson2025enhancing,Jhajj_JPllmasr_2025,kabir2023llm}. Unlike syntactic tree structures, KGs are directed graphs that can contain cycles and multiple paths between nodes (entities). In this preliminary work, we propose that surprise in KG reasoning corresponds to graph distance, where the KG serves as the agent's generative model. Entities that require shorter paths from context are unsurprising, whereas distant or disconnected entities are more surprising. This is unlike surprise-driven exploration in RL \citep{pathak_curiosity-driven_2017,rakotoaritina2025informationtheoretic}, where agents maximize surprise to explore, FEP agents minimize surprise by maintaining accurate generative models. Our work connects the FEP to practical KG systems through shortest-path distance, providing theoretical foundations for graph neural networks \citep{kipf_semi-supervised_2017, gangemi_modeling_2018, battaglia_relational_2018} and model-based reinforcement learning \citep{sutton_reinforcement_2018, verbelen_relationship_2020}.

\section{From Syntax to Semantics}

Murphy et al.\ \cite{murphy_natural_2024} quantified syntactic surprise via tree depth. We extend this to arbitrary directed graphs with cycles. Given a KG $\mathcal{G} = (\mathcal{E}, \mathcal{R}, \mathcal{T})$ with entities $\mathcal{E}$, relations $\mathcal{R}$, and triples $\mathcal{T} \subseteq \mathcal{E} \times \mathcal{R} \times \mathcal{E}$, geometric surprise is:

\begin{equation}
S_{\text{geo}}(e \mid C) = \begin{cases}
\displaystyle\min_{c \in C} d_{\mathcal{G}}(c, e) & \text{if path exists} \\[4pt]
\alpha & \text{otherwise}
\end{cases}
\label{eq:geo_surprise}
\end{equation}
where $d_{\mathcal{G}}(c, e)$ is the shortest directed path length from context $c \in C$ to entity $e$ (computed via BFS, Appendix~\ref{app:math}), and $\alpha$ is a hyperparameter penalizing disconnection. In our worked example, we set $\alpha = 5$; in general, $\alpha$ should exceed the graph's diameter (longest shortest-path distance) to ensure disconnected entities always have higher surprise than any connected entity. Combined with algorithmic complexity \citep{murphy_natural_2024}:
\begin{equation}
F(e \mid C) = S_{\text{geo}}(e \mid C) + \lambda K(\pi_{C \to e})
\label{eq:free_energy}
\end{equation}
where $K(\pi_{C \to e})$ is Kolmogorov complexity of the relation path, approximated 
via Lempel-Ziv compression, and $\lambda$ weights the components. For trees, this 
recovers Murphy's tree depth; for general graphs, it handles cycles naturally.

\textbf{Connection to FEP}: Under FEP, agents minimize $F = -\log P(o,s) - H[Q(s)]$ \citep{friston_free-energy_2010}. Interpreting the KG as the agent's generative model, we posit $-\log P(e \mid C) \propto d_{\mathcal{G}}(C,e)$: shorter distances indicate higher probability. Thus $S_{\text{geo}}$ implements the surprise term, while $K(\pi)$ approximates $H[Q(s)]$. Figure~\ref{fig:main} illustrates this with a political KG example (detailed calculations in Appendix~\ref{app:example}).

\begin{figure}[h!]
\centering
\includegraphics[width=\textwidth]{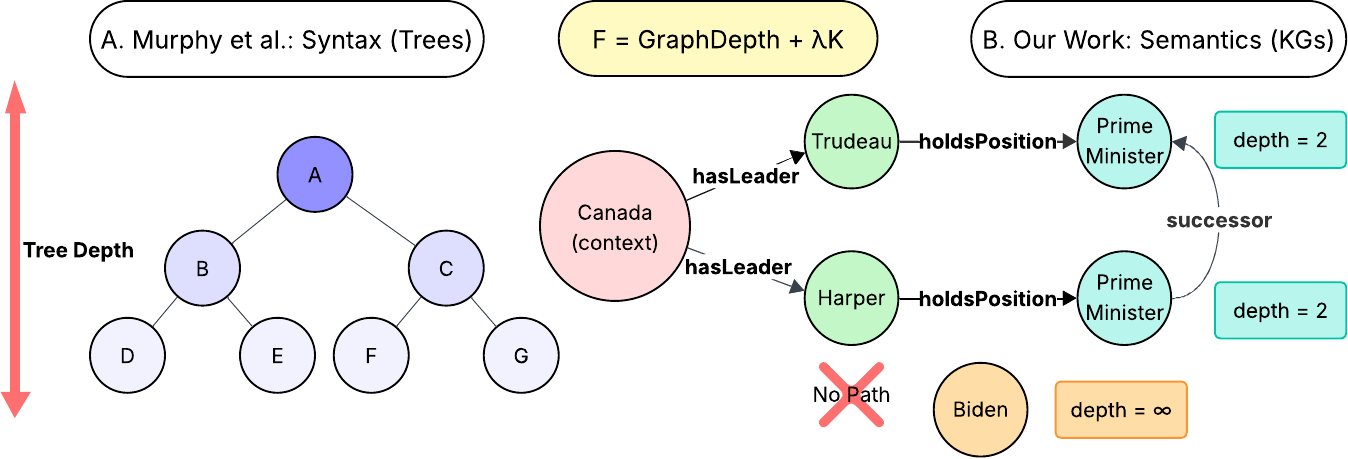}
\caption{\textbf{Extending surprise from trees to knowledge graphs.} Following standard KG design (e.g., Wikidata), we model ``Prime Minister'' as a position node. Given context ``Canada'', leaders (Trudeau, Harper) are at distance 1, the position node at distance 2, while disconnected entities (Biden) have distance $\infty$. The \texttt{successor} relation demonstrates cycle handling.}\label{fig:main}
\end{figure}

\section{Theoretical Justification}

Three principles justify the shortest-path distance: \textbf{(1) Proper generalization}: For trees, it recovers Murphy's tree depth. \textbf{(2) Least-action}: Shortest paths minimize cumulative cost, aligning with active inference where agents minimize expected free energy \citep{friston_active_2017}. \textbf{(3) Computational grounding}: In GNNs, $k$ message-passing iterations aggregate $k$-hop neighborhoods \citep{kipf_semi-supervised_2017, battaglia_relational_2018}; minimizing iterations minimizes distance and surprise. Cycles pose no issue: FEP accommodates circular causality \citep{friston_knowing_2015}, and BFS handles cycles via visited sets (Appendix~\ref{app:math}).

\section{Implications and Future Work}

This work-in-progress connects FEP from neuroscience to KG reasoning in AI systems. The presented framework offers practical implications: \textbf{(1) Entity grounding}: LLM-KG systems could rank candidate entity groundings by computing $S_{\text{geo}}$ via BFS from discourse context entities, preferring groundings with lower free energy \citep{jhajj_educational_2024, he_g-retriever:_2024}; \textbf{(2) KG embeddings}: embedding methods could preserve distance-based surprise structure \citep{bordes_translating_2013}; \textbf{(3) GNN architecture}: depth could be selected to balance computational cost against the surprise horizon needed for a task.

Future work includes empirical validation on benchmark KG datasets (FB15k-237 \cite{schlichtkrull2018modeling}, YAGO \cite{yago}), comparison with human semantic similarity judgments, integration with existing KG reasoning systems \citep{jhajj_educational_2024, deng_graphvis:_2024, he_g-retriever:_2024}, and extension to temporal KGs.
 
This work represents an early-stage exploration of applying FEP to knowledge graph reasoning. While we proposed the shortest-path distance as a principled formalization of surprise, other formulations may be more elegant or practical. 

We aim to present this contribution as an initial research direction rather than a definitive solution. We also encourage the community to develop complementary or improved approaches to connecting FEP principles with graph-based reasoning.

\begin{acknowledgments}

We acknowledge the support of the Natural Sciences and Engineering Research Council of Canada (NSERC), Alberta Innovates, Alberta Advanced Education, and Athabasca University, Canada. We would also like to thank the reviewers for their suggestions on how to improve this work.

\end{acknowledgments}


\section*{Declaration on Generative AI}

During the preparation of this work, the author(s) used Grammarly and Claude (Anthropic) for Grammar and spelling checks.

\newpage

\bibliography{sample-ceur}

\newpage
\appendix
\section{Worked Example: Free Energy Calculations}
\label{app:example}

We demonstrate free energy calculations using the Canadian Prime Minister 
knowledge graph from Figure~\ref{fig:main}.

\subsection{Scenario and Knowledge Graph}

Consider query ``Who is the Prime Minister?'' with context $C = \{\text{Canada}\}$. The knowledge graph contains:

\textbf{Entities}: $\mathcal{E} = \{\text{Canada}, \text{Trudeau}, \text{Harper}, 
\text{PrimeMinister}, \text{Biden}\}$

\textbf{Relations}: (Canada, \texttt{hasLeader}, Trudeau), (Canada, \texttt{hasLeader}, Harper), (Trudeau, \texttt{holdsPosition}, PrimeMinister), (Harper, \texttt{holdsPosition}, PrimeMinister),(Trudeau, \texttt{successor}, Harper), (Harper, \texttt{predecessor}, Trudeau)

The successor/predecessor relations form a cycle: Trudeau $\leftrightarrow$ Harper. Importantly, Biden has no directed path from Canada (separate subgraph).

\subsection{Computing Geometric Surprise}

Using BFS from Canada, we compute shortest directed paths:
\begin{itemize}
\item $d(\text{Canada}, \text{Trudeau}) = 1$ (direct via \texttt{hasLeader})
\item $d(\text{Canada}, \text{Harper}) = 1$ (direct via \texttt{hasLeader})
\item $d(\text{Canada}, \text{PrimeMinister}) = 2$ (via \texttt{hasLeader} then \texttt{holdsPosition})
\item $d(\text{Canada}, \text{Biden}) = \infty$ (no path)
\end{itemize}

Therefore: $S_{\text{geo}}(\text{Trudeau}) = S_{\text{geo}}(\text{Harper}) = 1$, $S_{\text{geo}}(\text{PrimeMinister}) = 2$, and $S_{\text{geo}}(\text{Biden}) = \alpha = 5$.

The cycle between Trudeau and Harper does not affect distances: BFS selects the shortest path (direct edge) and handles cycles via visited set (Appendix~\ref{app:math}).

\subsection{Computing Algorithmic Complexity}

For each grounding, we estimate Kolmogorov complexity via relation path patterns:

\textbf{Trudeau \& Harper}: Paths $\pi = [\texttt{hasLeader}]$ use frequent relations, yielding high compression (low $K(\pi)$).

\textbf{PrimeMinister node}: Path $\pi = [\texttt{hasLeader}, \texttt{holdsPosition}]$ uses standard role-modeling patterns, also yielding low $K(\pi)$.

\textbf{Biden}: No path from Canada. The grounding requires irregular cross-country reasoning not represented in the graph (high $K(\pi)$).

\subsection{Free Energy Results}

Combining components with $\lambda = 1$:

\begin{center}
\begin{tabular}{lccc}
\toprule
\textbf{Entity} & $S_{\text{geo}}$ & $K(\pi)$ & $F$ \\
\midrule
Trudeau & 1 & Low & $\sim$1.3 \\
Harper & 1 & Low & $\sim$1.3 \\
Biden & 5 & High & $\sim$5.5 \\
\bottomrule
\end{tabular}
\end{center}

\textbf{Interpretation}: Real groundings (Trudeau, Harper) exhibit low free energy: (1) short distance (1 hop), (2) regular relation patterns. The impossible grounding (Biden) exhibits high free energy: (1) disconnection (no path), (2) irregular pattern. The framework correctly identifies both Trudeau and Harper as plausible (both were Canadian PMs) while rejecting Biden (US president).

We focus on entity groundings (Trudeau, Harper, Biden) rather than the position node itself, as queries about leadership typically seek individuals rather than abstract roles. The PrimeMinister node, at distance 2, would have intermediate surprise ($S_{\text{geo}} = 2$, $F \approx 2.3$), but is not a direct answer to ``Who is the Prime Minister?'' This demonstrates how our framework naturally distinguishes between entities at different levels of abstraction in reified KG schemas.

This demonstrates three key properties: \textbf{(1)} cycles handled naturally, \textbf{(2)} multiple valid answers coexist with equal surprise, \textbf{(3)} disconnected entities correctly penalized.

\newpage
\section{Mathematical Details}
\label{app:math}

\subsection{Breadth-First Search Algorithm}

Given directed graph $\mathcal{G} = (\mathcal{E}, \mathcal{R}, \mathcal{T})$ and context $C \subseteq \mathcal{E}$, we compute $S_{\text{geo}}(e \mid C)$ via BFS:

\begin{algorithm}[H]
\caption{Compute Geometric Surprise}
\label{alg:bfs}
\begin{algorithmic}[1]
\REQUIRE Knowledge graph $\mathcal{G}$, context $C$, target entity $e$
\ENSURE Geometric surprise $S_{\text{geo}}(e \mid C)$
\STATE Initialize: $d(c) \leftarrow 0$ for all $c \in C$; $d(v) \leftarrow \infty$ for $v \notin C$
\STATE $Q \leftarrow C$ (queue), $V \leftarrow C$ (visited set)
\WHILE{$Q \neq \emptyset$}
    \STATE $u \leftarrow$ dequeue from $Q$
    \FOR{each outgoing edge $(u, r, v) \in \mathcal{T}$}
        \IF{$v \notin V$}
            \STATE $d(v) \leftarrow d(u) + 1$
            \STATE $V \leftarrow V \cup \{v\}$, enqueue $v$ to $Q$
        \ENDIF
    \ENDFOR
\ENDWHILE
\RETURN $d(e)$ if $d(e) < \infty$, else $\alpha$
\end{algorithmic}
\end{algorithm}

\textbf{Properties}: \textbf{(1)} \textit{Correctness}: BFS finds shortest paths in $O(|\mathcal{E}| + |\mathcal{T}|)$ time. \textbf{(2)} \textit{Cycle handling}: Visited set $V$ prevents re-visiting nodes, ensuring termination. \textbf{(3)} \textit{Directionality}: Only outgoing edges followed, respecting direction.

\subsection{Kolmogorov Complexity Approximation}

We approximate $K(\pi_{C \to e})$ via Lempel-Ziv compression: \textbf{(1)} Extract relation sequence $\pi = [r_1, \ldots, r_k]$ from shortest path. \textbf{(2)} Encode as string (e.g., ``pm|successor''). \textbf{(3)} Compress with LZ77. \textbf{(4)} Compute ratio $K(\pi) = \text{compressed}/\text{original}$.

\textbf{Interpretation}: Regular patterns (frequent relations, short sequences) achieve high compression (low $K$). Irregular patterns (rare relations, long sequences) achieve low compression (high $K$). This approximates Kolmogorov complexity, which is uncomputable \citep{li_introduction_2008}. Murphy et al.\ \citep{murphy_natural_2024} use the same approximation for syntactic patterns.

\subsection{Connection to Active Inference}

In active inference, agents minimize expected free energy $G(\pi)$ \citep{friston_active_2017, parr_active_2022}:
\begin{equation}
G(\pi) = \underbrace{D_{KL}[Q(o|\pi) \| P(o)]}_{\text{Pragmatic}} + 
\underbrace{\mathbb{E}_{Q(o|\pi)}[H[P(s|o)]]}_{\text{Epistemic}}
\end{equation}
balancing pragmatic value (exploitation) and epistemic value (exploration).

\textbf{Pragmatic value}: Entities at shorter distances are more likely: $P(\text{observe } e \mid C)$ increases as $S_{\text{geo}}$ decreases, making low-distance entities preferred for goal-directed actions.

\textbf{Epistemic value}: Entities at longer distances provide higher information gain: observing distant entities reduces uncertainty about unexplored graph regions, making high-distance entities preferred for exploration.

Our $S_{\text{geo}}$ implements pragmatic value: low surprise entities preferred for exploitation. Extensions could weight distance inversely for epistemic value, valuing high-surprise entities for exploration.

\end{document}